%% file: neurips_2026.tex
\documentclass{article}

\usepackage[preprint]{neurips_2026}

\usepackage[utf8]{inputenc}
\usepackage[T1]{fontenc}
\usepackage{hyperref}
\usepackage{url}
\usepackage{booktabs}
\usepackage{multirow}
\usepackage{array}
\usepackage{makecell}
\usepackage{algorithm}
\usepackage{algpseudocode}
\usepackage{amsmath}
\usepackage{amsfonts}
\usepackage{adjustbox}
\usepackage{graphicx}
\usepackage{xcolor}
\usepackage{tikz}
\usepackage{microtype}
\usepackage{caption}
\usepackage{enumitem}
\setlist[itemize]{leftmargin=1.2em, itemsep=0pt, parsep=0pt, topsep=1.5pt, partopsep=0pt}
\setlist[enumerate]{leftmargin=1.5em, itemsep=0pt, parsep=0pt, topsep=1.5pt, partopsep=0pt}
\usepackage{tcolorbox}
\tcbuselibrary{skins}
\usepackage{titlesec}
\titlespacing*{\paragraph}{0pt}{0.6ex plus 0.2ex minus 0.1ex}{0.8em}

\hypersetup{colorlinks=true, citecolor=blue, linkcolor=blue, urlcolor=blue}
\graphicspath{{img/}{img/}}

\title{Hybrid Retriever Evolution for Multimodal Document Reasoning Agents}

\author{
Bohan Yao$^{12}$ \quad Shruthan Radhakrishna$^{1}$ \quad Vikas Yadav$^{1}$\\
$^1$ServiceNow \quad $^2$University of Washington\\
\texttt{vikas.yadav@servicenow.com}
}

\begin{document}

\maketitle

\begin{abstract}
Different retrievers, including lexical, semantic, and multimodal approaches, provide highly complementary strengths for multimodal document understanding, yet most systems combine them through fixed pipelines that cannot adapt to the demands of individual reasoning steps. In this work, we ask whether retrieval orchestration itself can be \emph{learned} as part of the reasoning process. We introduce a failure-driven evolution framework in which a meta-agent autonomously discovers how a tool-using task agent should coordinate diverse retrievers during multi-step document question answering. The meta-agent analyzes incorrect reasoning trajectories, actively probes the same tool environment to diagnose root causes, and iteratively rewrites the task agent's instructions, turning retrieval from a fixed front-end stage into an adaptive, step-wise reasoning decision. The evolved agent learns \emph{when} to invoke each retriever, \emph{how} to combine them, and \emph{how} to compose evidence across modalities and pages. On MMLongBench-Doc and DocBench, the evolved agent achieves gains of up to $+19.6$ points over the unevolved baseline and consistently outperforms recent systems including MACT, MDocAgent, and SimpleDoc. Detailed retrieval analyses confirm that these improvements arise from adaptive routing and evidence composition rather than reliance on any hardcoded retrieval mode, and evolution dynamics reveal a progressive shift from narrow lexical behavior to rich multi-tool coordination. These findings establish autonomous multi-agent coordination as a promising paradigm for multimodal document reasoning.
\end{abstract}

\vspace{-2mm}

\input{latex/1_Introduction}
\input{latex/2_RelatedWork}
\input{latex/3_Method}
\input{latex/4_Experiments}
\input{latex/5_Results}
\input{latex/6_Analysis}
\input{latex/7_Conclusion}

\bibliographystyle{plainnat}
\bibliography{references}

\newpage
\appendix
\input{latex/9_Appendix}

\newpage
\input{checklist}

\end{document}

%% file: latex/1_Introduction.tex
\section{Introduction}

Multimodal documents such as scientific papers, financial reports, manuals, and business presentations communicate information through a rich mixture of text, tables, figures, charts, and layout structure. Answering questions over such documents is therefore not merely a retrieval problem: it is a long-horizon \emph{reasoning} problem that requires localizing evidence scattered across pages, integrating signals from multiple modalities, and reconciling partial or contradictory cues. Recent benchmarks such as MMLongBench-Doc and DocBench make this challenge starkly clear: many questions demand cross-page and cross-modal evidence aggregation, and even the strongest current systems still fall far short of robust performance \citep{ma2024mmlongbenchdoc,zou2024docbench}. These trends suggest that progress depends not only on stronger foundation models, but on fundamentally better mechanisms for evidence access and coordination.

A central difficulty is that no single retrieval paradigm is sufficient across all reasoning steps in these tasks. Lexical retrievers such as BM25 are effective for exact keyword matching and narrow grounding \citep{robertson2009probabilistic}; semantic text retrievers such as ColBERT improve conceptual matching and fine-grained relevance estimation \citep{khattab2020colbert}; multimodal retrievers such as ColPali capture visual and layout-dependent cues directly from document pages \citep{faysse2024colpali}. Prior work has shown the value of each retriever individually, and several recent systems combine text and image retrieval in hybrid pipelines \citep{han2025mdocagent,jain2025simpledoc}. However, most existing approaches still treat retrieval as a fixed front-end stage or rely on static fusion schemes, missing opportunities for dynamic adaptation when the most useful retriever changes from one reasoning step to the next, and when retrieved evidence must be filtered, compared, reconciled, or composed before it becomes useful.

In this paper, we investigate whether autonomous multi-agent systems can \emph{discover} how to utilize and combine diverse retrievers for multimodal document reasoning. Our key question is not which retriever is universally best, but whether an autonomous system can learn to use different retrievers meaningfully at different reasoning steps. Our framework pairs a tool-using task agent with a failure-driven meta-agent that improves it over time. The task agent answers document questions by deciding which tools to call, how to combine retrieved evidence, and when to stop. The meta-agent works offline: it inspects failed trajectories, actively explores the same tool environment to diagnose why the agent went wrong, and rewrites its instructions so that future trajectories use the toolbox more effectively. In this view, retrieval is no longer a static preprocessing module; it becomes an adaptive component of multi-step reasoning. This perspective complements prior work on improved retrievers and document agents: it asks whether autonomous multi-agent coordination can turn a diverse retrieval toolbox into an adaptive reasoning mechanism.

Our experiments provide evidence that failure-driven evolution can indeed discover stronger retrieval coordination strategies. With Gemini 3.1 Flash, the evolved agent improves from $42.4\%$ to $62.0\%$ on MMLongBench-Doc and from $73.4\%$ to $85.1\%$ on DocBench; with GPT-5-mini, from $40.7\%$ to $55.4\%$ and from $68.6\%$ to $79.3\%$. These gains are consistent across both benchmarks and backbones, and the evolved agent compares favorably with recent systems such as MACT, MDocAgent, and SimpleDoc. Importantly, the key message is not that our system achieves the highest numbers, but that it demonstrates a qualitatively different view of retrieval in multimodal document QA. The agent is not given a better retrieval pipeline; it discovers one through experience. Retrieval analyses confirm that ColPali is the strongest single retriever, yet substantial headroom remains from adaptive routing and fusion. Evolution analyses further reveal a non-monotonic but consistent shift from narrow early behavior toward richer multi-tool coordination. Together, the message is clear: strong multimodal document QA needs better retrieval \emph{decisions}, not just stronger standalone retrievers.


\begin{tcolorbox}[
  colback=green!7!white,
  colframe=green!40!black,
  boxrule=0.5pt,
  arc=4pt,
  left=6pt, right=6pt, top=5pt, bottom=5pt
]
Our main contributions are as follows:
\begin{itemize}
    \item We show that retrieval in multimodal document QA can be treated as an adaptive reasoning decision rather than a fixed preprocessing step, enabling an agent to decide when and how to use lexical, semantic, and multimodal retrievers across reasoning steps.

    \item We introduce a failure-driven evolution framework in which a meta-agent analyzes failed trajectories, probes the same tool environment, and rewrites the task agent's instructions to discover stronger retrieval-and-reasoning strategies.

    \item Across MMLongBench-Doc and DocBench, the evolved agent achieves consistent gains across two backbone models, improving by up to $+19.6$ points over the unevolved baseline without changing the retrievers, tools, or backbone.

    \item Our retrieval and evolution analyses show that the gains come from adaptive routing, evidence composition, and asymmetric complementarity among retrievers, rather than reliance on any single dominant retrieval mode.
\end{itemize}
\end{tcolorbox}

%% file: latex/2_RelatedWork.tex
\vspace{-2mm}
\section{Related Work}
\vspace{-1mm}
Recent progress in multimodal document question answering has been driven by two complementary trends: stronger retrieval for long, visually rich documents, and more capable agentic reasoning pipelines built on top of retrieved evidence. Our work sits at the intersection of both, asking whether retrieval orchestration itself can be learned as part of the reasoning process.

\paragraph{Retrieval for multimodal documents}
On the retrieval side, lexical methods such as BM25 remain effective for precise term matching and evidence localization \citep{robertson2009probabilistic}, while late-interaction dense retrievers such as ColBERT improve semantic matching through fine-grained token-level relevance estimation \citep{khattab2020colbert}. More recently, multimodal retrievers such as ColPali have demonstrated that embedding document pages as images captures layout, visual structure, and non-textual cues that text-only pipelines miss \citep{faysse2024colpali}. Together, these lines of work show that different retrieval paradigms offer distinct and complementary strengths for document understanding. Rather than proposing yet another retriever, we build on this complementarity and study how an agent can \emph{discover when each retriever is useful}.

\paragraph{Long-document multimodal QA}
Benchmarks such as DocVQA, InfographicVQA, and ChartQA established the importance of reasoning over visually structured documents \citep{appalaraju2021docvqa,mathew2022infographicvqa,masry2022chartqa}. More recent long-document benchmarks such as MMLongBench-Doc and DocBench push this further by emphasizing cross-page localization, multimodal evidence aggregation, and long-horizon reasoning over documents spanning dozens or hundreds of pages \citep{ma2024mmlongbenchdoc,zou2024docbench}. These benchmarks expose a core tension in current systems: the challenge is no longer merely to retrieve relevant context, but to retrieve the \emph{right kind} of evidence at the \emph{right step} of the reasoning process, often across multiple modalities and pages. This reveals a regime where retrieval can no longer be a one-shot preprocessing step; the system must iteratively gather and refine evidence as the reasoning state evolves.

\paragraph{Agentic document QA and multi-agent systems:}Recent document agents have moved beyond single-pass retrieval, but most still preserve fixed retrieval structure. MDocAgent uses specialized agents for text and image processing within a predefined multi-stage pipeline in which retrieval roles are assigned in advance \citep{han2025mdocagent}. SimpleDoc shows that a lightweight iterative framework with a dual-cue retriever can be effective, yet its retrieval mechanism is built around a fixed design rather than dynamically selecting among different retrievers \citep{jain2025simpledoc}. MACT demonstrates the value of multi-agent collaboration and test-time scaling for visual document understanding, but its main innovation lies in role-specialized reasoning rather than learning how retrieval should adapt across steps \citep{yu2025mact}. More broadly, LLM multi-agent frameworks such as AutoGen and AgentVerse show that decomposition and coordination improve complex tasks \citep{wu2023autogen,chen2024agentverse}; our setting is narrower, focusing on how document agents should \emph{route retrieval and compose evidence}. These methods show that decomposition, iteration, and collaboration matter, but stop short of treating retrieval orchestration as a learned reasoning problem. Our method is also related to approaches that improve LLM systems by optimizing prompts, modules, or execution strategies. DSPy and related work treat language-model pipelines as systems that can be compiled or optimized rather than hand-tuned \citep{khattab2023dspy}, while large language models have been used directly as optimizers over prompts and procedures \citep{yang2024largelanguagemodeloptimizers}. 

%% file: latex/3_Method.tex
\vspace{-1mm}
\section{Method}
\label{sec:method}
\vspace{-1mm}

\begin{figure}[t]
    \centering
    \includegraphics[width=\textwidth]{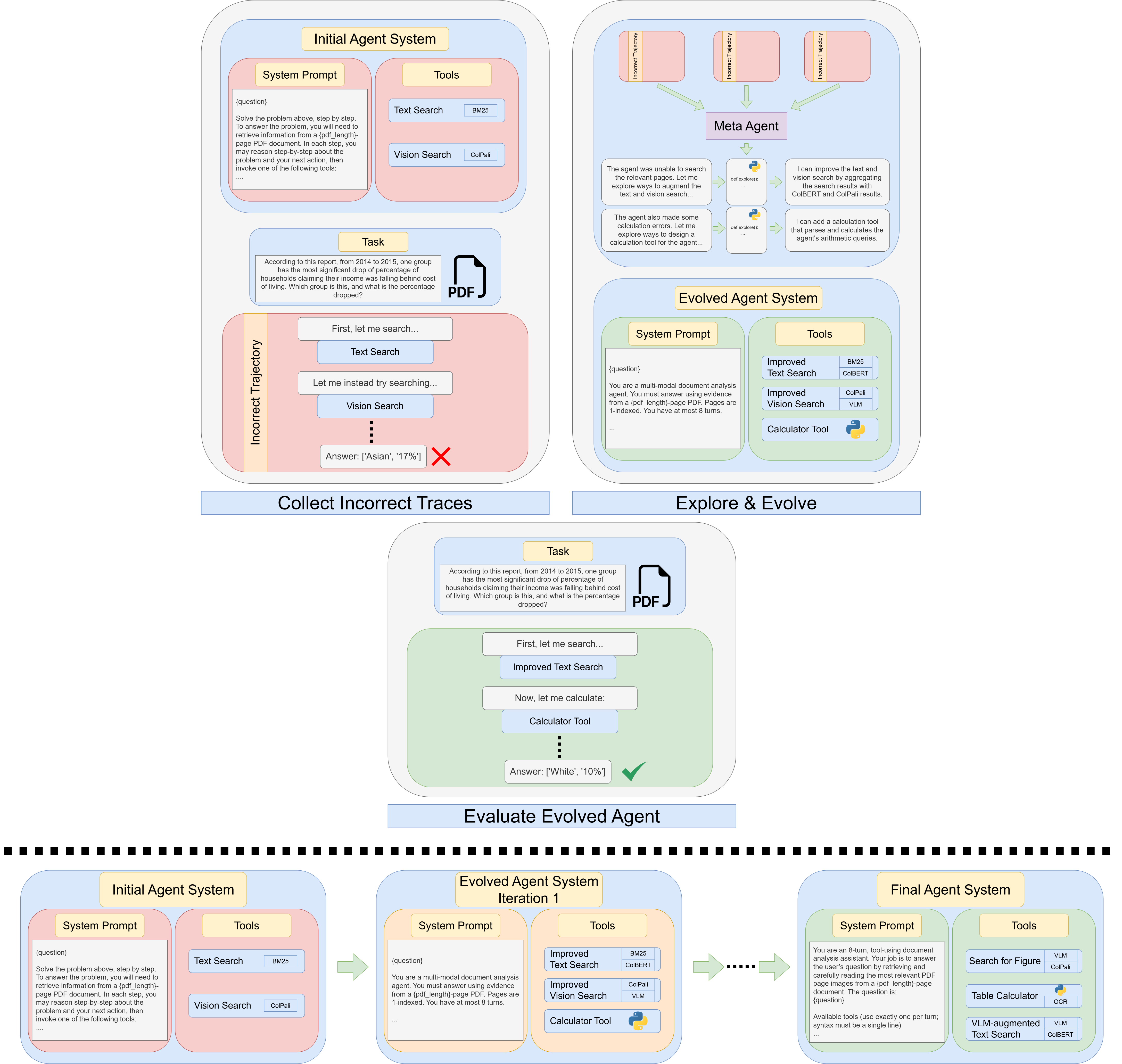}
    \caption{Overview of the failure-driven evolution framework for adaptive multimodal document reasoning. The \textbf{task agent} (left) answers document questions through iterative multi-step reasoning, deciding at each step which retrieval tool to invoke (lexical via BM25, semantic via ColBERT, multimodal via ColPali, or VLM-based retrieval) and how to compose, reconcile, and validate evidence drawn from different modalities and pages. When the task agent produces an incorrect answer, the \textbf{meta-agent} (right) receives the full failed trajectory together with the gold answer, then actively explores the same tool environment by generating and executing analysis code against BM25, ColBERT, ColPali, and the VLM tool. Through this interactive debugging, analogous to a human developer probing retrievers and inspecting returned evidence, the meta-agent diagnoses the root cause of the failure and proposes targeted updates to the task agent's system prompt ($\mathcal{P}_{sys}$) and tool-use parser ($\mathcal{P}_{tool}$). Proposed updates are accepted only if they strictly improve aggregate performance on a held-out validation batch, guarding against overfitting to individual failure cases. Over successive iterations, this process yields progressively stronger retrieval coordination and evidence composition strategies. At inference time, only the evolved task agent is deployed.}
    \vspace{-4mm}
    \label{fig:main_diagram}
\end{figure}

\begin{algorithm}[t]
\caption{\footnotesize Failure-driven prompt evolution. Each iteration: (1)~find a failure, (2)~actively probe the tool environment to diagnose it, (3)~propose prompt edits, (4)~keep edits only if they improve the validation batch.}
\vspace{-1mm}
\label{alg:prompt_evolution}
\begin{algorithmic}[1]
\Require Validation dataset $\mathcal{D}$, seed system prompt $\mathcal{P}_{sys}^{(0)}$, seed tool parser $\mathcal{P}_{tool}^{(0)}$
\Require Number of iterations $N$, exploration turns $T_{expl}$, batch size $B$
\Require Task agent model $\mathcal{M}_{agent}$, meta-agent model $\mathcal{M}_{evolve}$
\State $\mathcal{P}_{sys} \gets \mathcal{P}_{sys}^{(0)}$
\State $\mathcal{P}_{tool} \gets \mathcal{P}_{tool}^{(0)}$
\For{$i = 1 \dots N$}
    \State \textbf{Step 1: Failure discovery}
    \State $(q, a, d) \gets \text{None}$
    \For{each problem in $\mathcal{D}$}
        \State $\hat{a}, \mathcal{T} \gets \text{RunAgent}(\mathcal{M}_{agent}, \mathcal{P}_{sys}, \mathcal{P}_{tool}, \text{question}, \text{doc})$
        \State $\text{score} \gets \text{EvalScore}(\hat{a}, \text{answer})$
        \If{$\text{score} \neq 1.0$}
            \State $(q, a, d) \gets (\text{question}, \text{answer}, \text{doc})$
            \State \textbf{break}
        \EndIf
    \EndFor
    \State \textbf{Step 2: Interactive exploration}
    \State $\mathcal{H} \gets \{(q, a, d, \mathcal{P}_{sys}, \mathcal{P}_{tool}, \mathcal{T})\}$
    \For{$t = 1 \dots T_{expl}$}
        \State $C_t \gets \mathcal{M}_{evolve}(\mathcal{H})$
        \State $R_t \gets \text{Execute}(C_t)$
        \State $\mathcal{H} \gets \mathcal{H} \cup \{R_t\}$
    \EndFor
    \State \textbf{Step 3: Prompt evolution}
    \State $\hat{\mathcal{P}}_{sys}, \hat{\mathcal{P}}_{tool} \gets \mathcal{M}_{evolve}(\mathcal{H})$
    \State \textbf{Step 4: Evaluation and selection}
    \State Sample batch $\mathcal{B} \subset \mathcal{D}$ of size $B$
    \State $S_{old} \gets \sum_{j \in \mathcal{B}} \text{EvalScore}(\text{RunAgent}(\mathcal{M}_{agent}, \mathcal{P}_{sys}, \mathcal{P}_{tool}, q_j, d_j), a_j)$
    \State $S_{new} \gets \sum_{j \in \mathcal{B}} \text{EvalScore}(\text{RunAgent}(\mathcal{M}_{agent}, \hat{\mathcal{P}}_{sys}, \hat{\mathcal{P}}_{tool}, q_j, d_j), a_j)$
    \If{$S_{new} > S_{old}$}
        \State $\mathcal{P}_{sys} \gets \hat{\mathcal{P}}_{sys}$
        \State $\mathcal{P}_{tool} \gets \hat{\mathcal{P}}_{tool}$
    \EndIf
    \State $\mathcal{D} \gets \mathcal{D} \setminus \{(q, a, d)\}$
\EndFor
\State \Return $\mathcal{P}_{sys}, \mathcal{P}_{tool}$
\end{algorithmic}
\vspace{-1mm}
\end{algorithm}

Figure~\ref{fig:main_diagram} provides an overview of our approach. We introduce a failure-driven prompt evolution framework (Algorithm~\ref{alg:prompt_evolution}) in which a highly capable meta-agent ($\mathcal{M}_{evolve}$) diagnoses the failures of a task agent ($\mathcal{M}_{agent}$), actively explores the document environment to understand the context of each failure, and proposes systematic improvements to the agent's core instructions. The framework contains two complementary roles. The \emph{task agent} answers document questions by iteratively querying retrieval and auxiliary tools, deciding at each step which evidence to gather and how to compose it. The \emph{meta-agent} operates offline during evolution: it observes failed task-agent trajectories, actively investigates why they failed by probing the same tool environment with executable analysis code, and rewrites the task agent's instructions so that future trajectories use the retrieval toolbox more effectively. At inference time, only the evolved task agent is deployed.
\vspace{-1mm}
\subsection{Task Agent and Tool Environment}
The task agent operates over long multimodal documents by iteratively querying external tools and updating an internal scratchpad. We equip the agent with a diverse toolset: lexical retrieval with BM25, semantic retrieval with ColBERT, multimodal page retrieval with ColPali, a VLM-based page retrieval tool, and auxiliary utilities such as a calculator when explicit arithmetic is required. Crucially, the agent can call one tool or several in sequence, compare their outputs, impose additional conditions over retrieved evidence, and continue searching if the current evidence is incomplete or contradictory. The agent's behavior is dictated by two primary instruction blocks. The \emph{system prompt} $\mathcal{P}_{sys}$ defines the agent's overarching strategy: decompose the question, decide whether more retrieval is needed, reconcile intermediate evidence from different modalities, and answer only once the evidence is sufficient. The \emph{tool parser} $\mathcal{P}_{tool}$ dictates the syntax and logic for invoking external tools and specifies how retrieved outputs should be converted back into the agent's reasoning context. This separation is useful in practice because the system prompt encodes high-level reasoning behavior while the tool parser governs the mechanics of tool use.

\vspace{-2mm}
\subsection{Failure-Driven Evolution}
To improve the task agent, we introduce the prompt evolution procedure detailed in Algorithm~\ref{alg:prompt_evolution}. The evolution process is initialized with seed prompts and occurs over $N$ discrete iterations. Each iteration comprises four distinct phases: failure discovery, interactive exploration, prompt evolution, and empirical selection.

\newcommand{\passage}[1]{\vspace{0.4mm}\noindent\textbf{#1}}
\passage{Failure discovery:}
Rather than optimizing against generic instructions, we target concrete empirical failures. The current task agent is run over a validation pool until it produces an incorrect answer. We store the failed question, the gold answer, the associated document, and the full trajectory $\mathcal{T}$ of tool calls and intermediate reasoning that led to the mistake.

\passage{Interactive exploration:}
A failed trajectory alone is often not enough to diagnose why the agent broke. Because document QA requires nuanced understanding of document structure and visual layout, static analysis is often insufficient. The meta-agent therefore receives the failed trajectory together with the gold answer and is allowed to actively explore the same tool environment for $T_{expl}$ turns. During this phase, the meta-agent generates executable Python code that interfaces directly with the agent's exact toolset (BM25, ColBERT, ColPali, or the VLM retrieval tool) and inspects the returned text, page references, or bounding-box evidence. This interactive debugging mimics a human developer inspecting variables and executing test queries to understand why a system failed: the meta-agent does not guess blindly, it probes the environment.

\passage{Prompt evolution:}
Following the exploration phase, the meta-agent synthesizes its findings. It analyzes the gap between the agent's trajectory and the ground truth, conditioned on the newly retrieved document context, and proposes an updated system prompt $\hat{\mathcal{P}}_{sys}$ and tool parser $\hat{\mathcal{P}}_{tool}$. These edits can change how the task agent prioritizes retrievers, when it invokes multiple tools in conjunction, how it validates and reconciles evidence, and when it uses auxiliary computation.

\passage{Empirical selection:}
We do not accept edits just because they sound clever. To prevent catastrophic forgetting and ensure the proposed prompts generalize beyond the specific failure case, we rigorously evaluate each update before adopting it. We instantiate two parallel agents, the current agent and the candidate agent, and evaluate both on a randomly sampled validation batch. The update is kept if and only if the new agent achieves a strictly higher aggregate score. Finally, to ensure continuous exposure to novel challenges, the failure case utilized for the current iteration is dropped from the active dataset pool. We repeat this process over multiple iterations and use the strongest evolved checkpoint for final evaluation.

%% file: latex/4_Experiments.tex
\vspace{-2mm}
\section{Experiments}
\vspace{-2mm}

\newcommand{\passagen}[1]{\vspace{0.4mm}\noindent\textbf{#1}}

\passagen{Benchmarks and Metrics:}
We evaluate on two challenging multimodal document QA benchmarks that stress the long-horizon, cross-modal reasoning our framework addresses: \textbf{MMLongBench-Doc} and \textbf{DocBench} \citep{ma2024mmlongbenchdoc,zou2024docbench}. Both require evidence scattered across multiple pages and modalities, precisely the setting where adaptive retrieval coordination matters most. For end-to-end evaluation, we report the official benchmark QA score as a percentage. For retrieval-focused experiments, we report page-level precision, recall, and F1 against gold evidence pages. For oracle routing analysis, we additionally report results on the 853 answerable MMLongBench-Doc examples for which gold-evidence-based routing diagnostics are well defined.

\passagen{Models, Tools, and Baselines:}
Our task agent is evaluated with two LLM backbones in the main comparison: \textbf{Gemini 3.1 Flash} and \textbf{GPT-5-mini}. To verify that the observed retrieval trends generalize across reader architectures, the reader-side oracle study additionally includes \textbf{Qwen3-VL-8B} and \textbf{Ministral-3-14B-R}. Crucially, the tool environment is held \emph{fixed} across all agent variants. The retrieval pool contains BM25 for lexical search, ColBERT for semantic retrieval, ColPali for multimodal page retrieval, and a VLM-based retriever; the evolved system additionally invokes auxiliary tools when explicit arithmetic is required. By holding tools constant, we isolate the effect of a better retrieval-and-reasoning \emph{policy} from the confound of giving the evolved agent a richer environment. We compare against baselines spanning simple prompting to recent agentic systems. \textbf{CoT} answers directly with chain-of-thought prompting without retrieval. \textbf{File API} gives the base model direct access to the full document without our adaptive controller. \textbf{MACT}, \textbf{MDocAgent}, and \textbf{SimpleDoc} are recent agentic or retrieval-augmented document QA systems \citep{yu2025mact,han2025mdocagent,jain2025simpledoc}. \textbf{Baseline Agent (Ours)} is the unevolved task agent with the same backbone and tools as our final system. \textbf{Evolved Agent (Ours)} uses the same backbone and tools, but with prompts and tool-use policy discovered through the evolution procedure. We also evaluate ablation variants that remove ground-truth access during evolution, substitute a weaker meta-agent, or disable interactive exploration, to isolate each component's contribution.

\passagen{Evolution and Evaluation Protocol:}
We initialize the system with a seed system prompt and tool parser, then run the failure-driven evolution loop from Algorithm~\ref{alg:prompt_evolution}. In each iteration, the meta-agent identifies an incorrect validation trajectory, actively probes the tool environment to diagnose the failure, proposes targeted edits to the task agent's instructions, and keeps them only if they improve a validation batch. This produces progressively refined task agents rather than a single one-shot prompt edit. For the main benchmark comparison, we report the strongest evolved checkpoint. For the evolution-dynamics study, we report all 20 checkpoints to reveal how both performance and tool-use behavior change over time. For the retriever and oracle studies, we use controlled retrieval setups to separate three effects: the quality of individual retrievers, the headroom from better routing and fusion, and the extent to which stronger retrieval translates into stronger downstream QA. Oracle settings serve only as upper-bound diagnostics and are not realizable at test time.

\vspace{-1mm}

%% file: latex/5_Results.tex
\vspace{-2mm}
\section{Results}

\begin{table*}[t]
\centering
\resizebox{0.95\textwidth}{!}{%
\begin{tabular}{>{\raggedright\arraybackslash}m{1.1cm} l c c c}
\toprule
\textbf{Model} & \textbf{Method} & \textbf{Evolved?}$^\dagger$ & \textbf{MMLongBench-Doc} & \textbf{DocBench} \\
\midrule
\multirow{10}{*}{\rotatebox[origin=c]{90}{\normalsize Gemini-3.1-Flash}}
& CoT & \textcolor{red}{$\times$} & 40.4\% & 66.1\% \\
& File API & \textcolor{red}{$\times$} & 45.4\% & 71.7\% \\
& MACT & \textcolor{red}{$\times$} & 54.4\% & 79.7\% \\
& MDocAgent & \textcolor{red}{$\times$} & 51.2\% & 78.6\% \\
& SimpleDoc & \textcolor{red}{$\times$} & 56.6\% & 82.2\% \\
\cmidrule(lr){2-5}
& \textbf{Baseline Agent (Ours)} & \textcolor{red}{$\times$} & 42.4\% & 73.4\% \\
& \textbf{Evolved Agent, No Ground Truth (Ours)} & \textcolor{green}{$\checkmark$} & 58.4\% & 84.0\% \\
& \textbf{Evolved Agent, GPT-5-minimal Meta Agent (Ours)} & \textcolor{green}{$\checkmark$} & 56.1\% & 79.8\% \\
& \textbf{Evolved Agent, No Exploration (Ours)} & \textcolor{green}{$\checkmark$} & 57.4\% & 77.3\% \\
& \textbf{Evolved Agent (Ours)} & \textcolor{green}{$\checkmark$} & \textbf{62.0\%} & \textbf{85.1\%} \\
\midrule
\multirow{10}{*}{\rotatebox[origin=c]{90}{\normalsize GPT-5-mini}}
& CoT & \textcolor{red}{$\times$} & 36.2\% & 57.4\% \\
& File API & \textcolor{red}{$\times$} & 44.6\% & 72.1\% \\
& MACT & \textcolor{red}{$\times$} & 47.4\% & 73.3\% \\
& MDocAgent & \textcolor{red}{$\times$} & 44.0\% & 70.1\% \\
& SimpleDoc & \textcolor{red}{$\times$} & 51.3\% & 74.8\% \\
\cmidrule(lr){2-5}
& \textbf{Baseline Agent (Ours)} & \textcolor{red}{$\times$} & 40.7\% & 68.6\% \\
& \textbf{Evolved Agent, No Ground Truth (Ours)} & \textcolor{green}{$\checkmark$} & 54.9\% & 75.6\% \\
& \textbf{Evolved Agent, GPT-5-minimal Meta Agent (Ours)} & \textcolor{green}{$\checkmark$} & 52.8\% & 75.6\% \\
& \textbf{Evolved Agent, No Exploration (Ours)} & \textcolor{green}{$\checkmark$} & 55.5\% & 74.8\% \\
& \textbf{Evolved Agent (Ours)} & \textcolor{green}{$\checkmark$} & \textbf{55.4\%} & \textbf{79.3\%} \\
\bottomrule
\end{tabular}%
}
\caption{End-to-end QA results on MMLongBench-Doc and DocBench. The baseline and evolved agents share the same backbone and tool environment; evolution changes only the retrieval-and-reasoning \emph{policy}. Ablation variants isolate ground-truth supervision, meta-agent capability, and interactive exploration. The evolved agent achieves gains of $+19.6$ and $+11.7$ points with Gemini 3.1 Flash and $+14.7$ and $+10.7$ with GPT-5-mini, consistently outperforming all baselines. $^\dagger$Evolved? indicates that the agent's instructions were refined through evolution; the tool environment is held fixed across all rows.}
\vspace{-2mm}
\label{tab:main_results}
\end{table*}

\vspace{-2mm}

\paragraph{End-to-end performance:}We evaluate the framework end-to-end, directly testing the central question: can retrieval orchestration be \emph{learned} as part of the reasoning process? Table~\ref{tab:main_results} provides a clear affirmative answer. With Gemini 3.1 Flash, evolution delivers gains of $19.6$ and $11.7$ points on MMLongBench-Doc and DocBench; with GPT-5-mini, the gains are $14.7$ and $10.7$ points. These improvements are notable because the baseline and evolved agents share the same backbone and tools; only the retrieval-and-reasoning \emph{policy} changes. The unevolved agent can trail specialized systems such as SimpleDoc or MACT; the performance jump comes \emph{only} after the meta-agent rewrites how the task agent searches, validates, and composes evidence. This reveals why the multi-agent approach offers something qualitatively different from fixed retrieval pipelines: the agent does not benefit from better tools, but from discovering \emph{when} to use each retriever and \emph{how} to compose their outputs. This is what the framework was designed to isolate (Section~\ref{sec:method}): stronger document QA from better retrieval decisions, not richer tools.

\paragraph{Ablation analysis}:Table~\ref{tab:main_results} further reports three ablation variants that isolate each design decision in the evolution framework, testing which components of the meta-agent's four-phase process are essential. \textbf{No Ground Truth} withholds gold answers during evolution, forcing the meta-agent to identify failures without oracle supervision; performance drops modestly ($58.4\%$ vs.\ $62.0\%$ on MMLongBench-Doc with Gemini), suggesting the framework remains effective even under weaker supervision. \textbf{GPT-5-minimal Meta Agent} substitutes a smaller meta-agent; the consistent drop across both benchmarks and backbones confirms that meta-agent capability meaningfully influences evolution quality, consistent with the view that diagnosing retrieval failures in complex documents requires strong reasoning. \textbf{No Exploration} removes the interactive tool-probing phase (Step~2 of Algorithm~\ref{alg:prompt_evolution}), the component that lets the meta-agent actively query the same tools as the task agent; the gap is most pronounced on DocBench ($77.3\%$ vs.\ $85.1\%$ with Gemini), underscoring that active exploration is especially valuable for visually rich documents where static trajectory analysis is insufficient. Together, these confirm that each component contributes to the final result.

\paragraph{Retrieval quality:}
\label{sec:retrieval_results}
\label{sec:reader_oracle_results}
Table~\ref{tab:retrieval_qa_results} consolidates retriever comparison and downstream QA: the left columns report page-level evidence recovery (precision, recall, F1), while the right columns report QA accuracy with four reader models. BM25 and VLM retrieval provide useful recall but limited F1; ColBERT improves through semantic matching; ColPali performs substantially better than single-retriever baselines, highlighting the importance of multimodal page representations. These patterns confirm that retriever strength alone cannot explain the evolved agent's gains; the diversity across retrieval paradigms creates the opportunity for adaptive orchestration that evolution exploits.

\begin{table*}[t]
\centering
\small
\renewcommand{\arraystretch}{1.15}
\setlength{\tabcolsep}{4.5pt}
\resizebox{\textwidth}{!}{%
\begin{tabular}{l ccc cccc}
\toprule
 & \multicolumn{3}{c}{\textbf{Evidence Recovery}} & \multicolumn{4}{c}{\textbf{End-to-End QA (MMLongBench-Doc)}} \\
\cmidrule(lr){2-4} \cmidrule(lr){5-8}
\textbf{Method} & \textbf{Prec.} & \textbf{Recall} & \textbf{F1} & \textbf{Qwen3-VL-8B} & \textbf{Ministral-3-14B-R} & \textbf{GPT-5-mini} & \textbf{Gemini 3.1 Flash} \\
\midrule
VLM retrieval (Qwen2-VL-2B)    & 17.5 & 48.2 & 23.3 & 46.8 & 42.1 & 47.8 & 47.2 \\
Lexical retrieval (BM25)       & 16.9 & 55.6 & 24.3 & 35.7 & 28.6 & 36.3 & 34.6 \\
Semantic retrieval (ColBERT)   & 18.3 & 59.4 & 26.1 & 47.0 & 37.8 & 44.3 & 45.2 \\
Multimodal retrieval (ColPali) & 24.1 & 77.2 & 34.3 & 52.8 & 43.0 & 49.6 & 52.7 \\
\midrule
Oracle 1$^{\dagger}$ (best retriever per query)        & 25.9 & 81.9 & 36.7 & 53.7 & 42.9 & 56.2 & 44.8 \\
Oracle 2$^{\dagger}$ (ColBERT + ColPali, top-$10$)     & 18.1 & 82.9 & 27.8 & 59.2 & 51.5 & 56.2 & 56.7 \\
Oracle 3a (gold-only fused evidence)                   & 90.7 & 82.9 & 85.3 & 40.4 & 38.0 & 40.1 & 34.8 \\
Oracle 3b (gold-prioritized fusion + fill)             & 32.0 & 93.4 & 43.9 & 57.6 & 46.7 & 53.1 & 55.8 \\
\midrule
Gold ceiling                                            & 100.0 & 100.0 & 100.0 & 66.7 & 64.1 & 65.8 & 60.2 \\
\bottomrule
\end{tabular}%
}
\caption{\scriptsize Retrieval quality and end-to-end QA on MMLongBench-Doc. \emph{Left:} page-level evidence recovery. \emph{Right:} QA accuracy with four reader models. ColPali is the strongest single retriever, but oracle routing and fusion reveal substantial headroom from adaptive selection and evidence composition. $^{\dagger}$Oracle 1 and Oracle 2 are diagnostic upper bounds, not realizable test-time systems.}
\vspace{-4mm}
\label{tab:retrieval_qa_results}
\end{table*}

\paragraph{Oracle headroom.}
The oracle configurations reveal significant untapped headroom beyond any single retriever. Oracle~1 (per-query best-retriever selection) improves over ColPali, confirming retriever complementarity even when one retriever dominates on average. Oracle~2 (union of ColBERT and ColPali top-$10$) reaches high recall but lower precision, showing that naive expansion increases coverage at the cost of noise. Oracle~3a (gold-only filtering) performs \emph{below the ColPali baseline}: overly restrictive filtering removes helpful context the reader needs to reason correctly. Oracle~3b (gold-prioritized with ColPali backfill) approaches the gold ceiling most closely, showing that successful evidence composition requires both page prioritization \emph{and} retention of useful surrounding context. The QA columns confirm these patterns carry through to end-task performance across all four readers. Taken together, the key opportunity lies not in choosing a better retriever, but in learning how to \emph{select the right retriever at each step and compose their outputs} into coherent evidence. This is precisely the capability the evolved multi-agent system discovers through failure-driven evolution.
\vspace{-2mm}

%% file: latex/6_Analysis.tex
\section{Analysis}
\vspace{-1mm}

We probe \emph{why} the evolved agent improves through two main lenses: evolution dynamics (\S\ref{sec:evolution_dynamics}) and failure complementarity (\S\ref{sec:failure_complementarity}). A detailed oracle routing analysis is provided in Appendix~\ref{sec:oracle_analysis}.
\vspace{-2mm}
\subsection{Evolution Dynamics Across Iterations}
\label{sec:evolution_dynamics}

\begin{figure}
    \centering
    \resizebox{!}{0.9\height}{%
        \includegraphics[width=\textwidth]{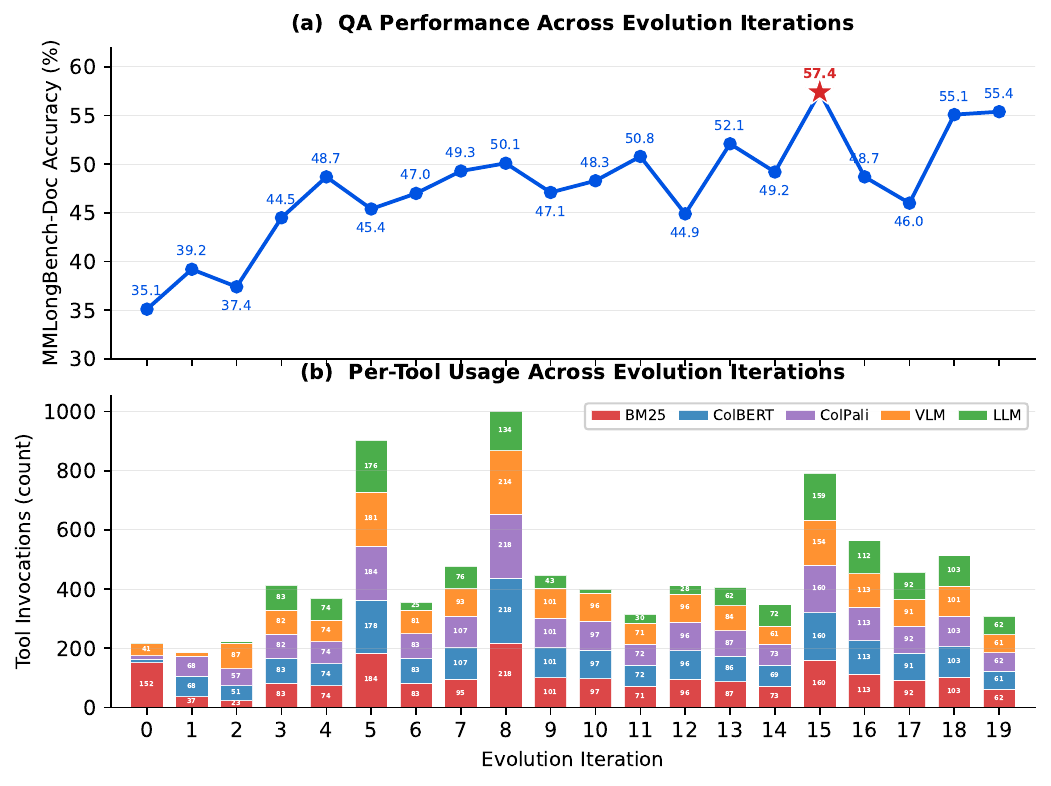}
    }
    \vspace{-3mm}
    \caption{Evolution dynamics across 20 iterations. \textbf{(a)}~QA accuracy improves non-monotonically from $35.1$ to a peak of $57.4$ (iteration~15), settling at $55.4$. \textbf{(b)}~Stacked tool invocation counts show a shift from narrow BM25 dominance (iteration~0: $152$ of $218$ calls) toward balanced multi-tool coordination. High-performing checkpoints do not correspond to any single dominant tool, confirming that gains arise from better orchestration.}
    \label{fig:evolution_dynamics}
    \vspace{-6mm}
\end{figure}

Does the evolution procedure from Section~\ref{sec:method} produce genuinely different retrieval policies, or merely surface-level prompt variation? Figure~\ref{fig:evolution_dynamics} answers this question by tracking both QA performance and per-tool invocation shares across all 20 iterations. Performance rises from $35.1$ at iteration~$0$ to a peak of $57.4$ at iteration~$15$, settling at $55.4$ by iteration~$19$. The trajectory is non-monotonic because each update targets a specific failure, but the overall trend reflects exploration and consolidation. Crucially, the tool-usage panel shows that high-performing checkpoints do \emph{not} achieve gains by favoring one tool: early lexical dominance (BM25: $152$ of $218$ calls at iteration~$0$) gives way to balanced multi-tool coordination. The benefit lies in learning \emph{when} to use each retriever, consistent with the complementarity pattern from Table~\ref{tab:main_results} and the oracle study (Appendix~\ref{sec:oracle_analysis}).
\vspace{-2mm}
\subsection{Failure Complementarity Across Retrievers}
\label{sec:failure_complementarity}

\begin{figure*}[t]
\begin{minipage}[t]{0.34\textwidth}
\vspace{1.15cm}
\centering
\scriptsize
\renewcommand{\arraystretch}{1.1}
\setlength{\tabcolsep}{3pt}
\begin{tabular}{lccccc}
\toprule
\textbf{Method} & \textbf{P} & \textbf{R} & \textbf{F1} & \textbf{Hit} & \textbf{Miss} \\
\midrule
VLM-all      & 17.5 & 48.2 & 23.3 & 56.4 & 43.6 \\
BM25         & 16.9 & 55.6 & 24.3 & 66.6 & 33.4 \\
ColBERT      & 18.3 & 59.4 & 26.1 & 71.7 & 28.3 \\
ColPali      & 24.1 & 77.2 & 34.3 & \textbf{86.1} & \textbf{13.9} \\
Agent-VLM    & \textbf{37.8} & 61.5 & \textbf{42.7} & 70.1 & 29.9 \\
Agent-LLM    & 28.3 & 44.8 & 31.4 & 53.0 & 47.0 \\
Agent-BM25s  & 14.0 & 46.2 & 20.2 & 54.8 & 45.2 \\
\bottomrule
\end{tabular}
\captionof{table}{Top-$5$ retrieval on 854 answerable MMLongBench-Doc questions. \textsc{ColPali} leads in coverage ($86.1\%$ hit rate); \textsc{Agent-VLM} leads in precision and F1.}
\label{tab:retrieval_hitmiss}
\end{minipage}
\hfill
\begin{minipage}[t]{0.64\textwidth}
\vspace{0pt}
\centering
\definecolor{wmblue}{HTML}{0053E2}
\resizebox{\linewidth}{!}{%
\begin{tikzpicture}[font=\small]
  \foreach \name [count=\ci from 0] in
    {VLM-all, BM25, ColBERT, ColPali, Ag-VLM, Ag-LLM, Ag-BM25s} {
    \node[rotate=45, anchor=south west, font=\footnotesize]
      at ({\ci*1.28+0.64}, 0.15) {\name};
  }

  \foreach \name [count=\ri from 0] in
    {VLM-all, BM25, ColBERT, ColPali, Ag-VLM, Ag-LLM, Ag-BM25s} {
    \node[anchor=east, font=\footnotesize]
      at (-0.1, {-\ri*1.0-0.5}) {\name};
  }

  \foreach \ri/\ci/\val in {
    0/0/-1,  0/1/57.0,  0/2/65.3,  0/3/76.9,  0/4/55.4,  0/5/37.1,  0/6/47.0,
    1/0/43.9, 1/1/-1,   1/2/41.4,  1/3/66.3,  1/4/58.9,  1/5/38.9,  1/6/13.0,
    2/0/46.7, 2/1/31.0,  2/2/-1,   2/3/67.4,  2/4/59.5,  2/5/40.9,  2/6/33.1,
    3/0/27.7, 3/1/19.3,  3/2/33.6,  3/3/-1,   3/4/37.8,  3/5/32.8,  3/6/20.2,
    4/0/34.9, 4/1/54.1,  4/2/61.6,  4/3/71.0,  4/4/-1,   4/5/34.9,  4/6/51.0,
    5/0/41.6, 5/1/56.6,  5/2/64.3,  5/3/80.0,  5/4/58.6,  5/5/-1,   5/6/33.4,
    6/0/49.0, 6/1/35.8,  6/2/58.0,  6/3/75.4,  6/4/67.6,  6/5/30.8,  6/6/-1%
  } {
    \pgfmathtruncatemacro{\isDiag}{\val < 0 ? 1 : 0}
    \ifnum\isDiag=1
      \fill[gray!15] ({\ci*1.28}, {-\ri*1.0}) rectangle ++(1.28, -1.0);
      \node[font=\footnotesize, text=gray!60]
        at ({\ci*1.28+0.64}, {-\ri*1.0-0.5}) {--};
    \else
      \pgfmathsetmacro{\pct}{max(0, min(100, (\val - 13.0) / (80.0 - 13.0) * 100))}
      \fill[wmblue!\pct!white]
        ({\ci*1.28}, {-\ri*1.0}) rectangle ++(1.28, -1.0);
      \pgfmathtruncatemacro{\useWhite}{\pct > 55 ? 1 : 0}
      \ifnum\useWhite=1
        \node[font=\footnotesize, text=white]
          at ({\ci*1.28+0.64}, {-\ri*1.0-0.5}) {\pgfmathprintnumber[fixed,precision=1]{\val}};
      \else
        \node[font=\footnotesize]
          at ({\ci*1.28+0.64}, {-\ri*1.0-0.5}) {\pgfmathprintnumber[fixed,precision=1]{\val}};
      \fi
    \fi
    \draw[gray!30] ({\ci*1.28}, {-\ri*1.0}) rectangle ++(1.28, -1.0);
  }

  \pgfmathsetmacro{\barX}{7*1.28 + 0.5}
  \foreach \s in {0,1,...,69} {
    \pgfmathsetmacro{\intensity}{\s / 69 * 100}
    \fill[wmblue!\intensity!white]
      (\barX, {-\s*7.0/70}) rectangle ++(0.35, {-7.0/70});
  }
  \draw[gray!50] (\barX, 0) rectangle ++(0.35, -7.0);
  \node[anchor=west, font=\scriptsize] at ({\barX+0.45}, 0)     {80\%};
  \node[anchor=west, font=\scriptsize] at ({\barX+0.45}, -3.5)  {47\%};
  \node[anchor=west, font=\scriptsize] at ({\barX+0.45}, -7.0)  {13\%};
  \node[anchor=south, font=\scriptsize, rotate=90]
    at ({\barX+1.0}, -3.5) {Rescue rate (\%)};

  \node[anchor=north, font=\footnotesize] at ({7*1.28/2}, -7.35)
    {Rescuer (method B)};
  \node[anchor=south, font=\footnotesize, rotate=90] at (-1.9, -3.5)
    {Failed method (A)};
\end{tikzpicture}%
}
\captionof{figure}{Rescue-rate heatmap. Each cell reports the percentage of method~A's misses (zero gold pages in top-$5$) that method~B rescues. \textsc{ColPali} (fourth column) rescues $66$--$80\%$ of every other method's failures. When ColPali itself fails, only Agent-VLM provides meaningful rescue ($37.8\%$).}
\label{fig:rescue_heatmap}
\end{minipage}
\vspace{-7mm}
\end{figure*}

Table~\ref{tab:retrieval_hitmiss} reports top-$5$ retrieval quality on 854 answerable questions. \textsc{ColPali} achieves the best recall and hit rate ($86.1\%$), while \textsc{Agent-VLM} leads on precision and F1, revealing a natural division between broad coverage and precise retrieval. The rescue-rate heatmap (Figure~\ref{fig:rescue_heatmap}) makes this complementarity concrete. Notably, \textsc{ColPali} is the best rescuer for \emph{every} other method ($66$--$80\%$ of misses rescued), explaining its standalone dominance throughout the paper. When \textsc{ColPali} itself fails, \textsc{Agent-VLM} is the only meaningful rescuer ($37.8\%$); since \textsc{ColPali} misses only $13.9\%$ of the answerable set, roughly one quarter of its failures are unrescuable, identifying a hard residual subset. Meanwhile, BM25 and \textsc{Agent-BM25s} overlap heavily (rescue rate $13.0\%$, the matrix minimum), whereas \textsc{ColBERT} and \textsc{Agent-VLM} form the most complementary pair ($\sim60.5\%$ symmetric rescue rate), confirming that semantic text and agent-guided visual retrieval capture different evidence. Together, the analyses converge: routing headroom is real and structured, the agent progressively discovers it through evolution, and it lives in asymmetric rescue relationships across retrieval modes.
\vspace{-3mm}

%% file: latex/7_Conclusion.tex
\vspace{-2mm}
\section{Conclusion}
\vspace{-2mm}

We opened this paper by asking whether retrieval in multimodal document QA should remain a fixed front-end stage or become an adaptive part of reasoning itself. The evidence points emphatically to the latter. By pairing a tool-using task agent with a failure-driven meta-agent that diagnoses incorrect trajectories, explores the document tool environment, and iteratively rewrites the task agent's instructions, we show that this process discovers stronger retrieval coordination strategies, yielding substantial gains on MMLongBench-Doc and DocBench across multiple LLM backbones. The supporting analyses reinforce this from multiple directions: ColPali is the strongest single retriever, yet oracle experiments (Appendix~\ref{sec:oracle_analysis}) reveal clear headroom from adaptive routing; evolution transforms the agent from narrow lexical behavior toward rich multi-tool coordination; and failure-complementarity analysis confirms that this headroom lies in structured rescue relationships across retrieval modes. Crucially, these gains arise from stronger retrieval policies, not superficial prompt variation. More broadly, instead of improving a fixed retrieval pipeline, we investigate whether retrieval usage itself can be \emph{evolved} as part of reasoning. Potential applications span scientific, legal, and enterprise domains where navigating long, multimodal reports reliably is critical. Deployment carries risks: incorrect evidence composition could mislead users on sensitive documents, and more capable agents may encourage over-reliance on opaque systems. We hope this work helps shift multimodal document QA toward systems in which accessing and composing evidence is a deliberate, step-wise reasoning decision rather than a predetermined pipeline.

%% file: latex/9_Appendix.tex
\section{Limitations}
\label{sec:limitations}

While the failure-driven evolution framework produces substantial gains, several limitations deserve acknowledgment.

\paragraph{Benchmark scope.}
We evaluate on two benchmarks, MMLongBench-Doc and DocBench, that emphasize long, multimodal documents. These are among the most challenging current benchmarks, and we further test across two LLM backbones (Gemini 3.1 Flash and GPT-5-mini) to verify that the gains are not model-specific. That said, generalization to other document distributions, such as scanned historical archives, heavily templated forms, or multilingual documents, remains untested. The retrieval strategies discovered by evolution may be specific to the kinds of documents and questions represented here. Extending evaluation to a broader set of document types and languages is a natural next step.

\paragraph{Computational cost.}
The evolution loop could be compute-intensive: each of the $N=20$ iterations involves running the task agent to discover a failure, executing $T_{\text{expl}}$ exploration turns with the meta-agent, and evaluating both the old and new agent on a validation batch. Importantly, the empirical selection step prevents wasted computation by discarding updates that do not improve validation performance, and the evolved agent at inference time adds no overhead beyond the unevolved baseline since only the refined prompt is deployed. The remaining gap is the absence of detailed runtime and cost breakdowns, which limits practitioners' ability to assess practical trade-offs. A systematic study of how performance scales with fewer iterations or smaller exploration budgets, and whether evolved policies can be distilled into cheaper forms, would make the approach more practical for resource-constrained settings.

\paragraph{Interpretability of evolved policies.}
The meta-agent rewrites the task agent's system prompt and tool parser in natural language, but the resulting policies are not easily interpretable as explicit retrieval-routing rules. Our evolution-dynamics and failure-complementarity analyses (Sections~\ref{sec:evolution_dynamics} and~\ref{sec:failure_complementarity}) provide indirect interpretability by revealing which tools the agent learns to favor and how rescue relationships shift, but they do not explain \emph{why} a specific retriever is chosen at a specific reasoning step. Distilling evolved prompts into explicit, inspectable routing policies, for example through learned decision trees or retriever-selection classifiers, is a promising direction that would make the approach both more transparent and more transferable.

\section{Oracle Analysis of Retriever Routing and Fusion}
\label{sec:oracle_analysis}

\begin{table*}[t]
\centering
\small
\renewcommand{\arraystretch}{1.15}
\setlength{\tabcolsep}{6pt}
\resizebox{\textwidth}{!}{%
\begin{tabular}{l c c c c c c c c c c c c c c}
\toprule
\textbf{Reader} & \textbf{N} & \textbf{ColBERT} & \textbf{ColPali} & \textbf{Best} & \textbf{Oracle 1} & \textbf{Oracle 2} & \textbf{Oracle 3a} & \textbf{Oracle 3b} & \textbf{Gold} & \textbf{O1} & \textbf{O2} & \textbf{O3a} & \textbf{O3b} & \textbf{O3b $\rightarrow$ Gold} \\
 & \textbf{ans.} &  &  & \textbf{baseline} &  &  &  &  & \textbf{ceiling} & \textbf{gain} & \textbf{gain} & \textbf{gain} & \textbf{gain} & \textbf{gap} \\
\midrule
Qwen3-VL-8B        & 853 & 43.1 & 52.1 & 52.1 & 53.7 & 57.1 & 51.2 & \textbf{58.6} & 59.2 & 1.6 & 5.0 & -1.0 & 6.5 & 0.6 \\
Ministral-3-14B-R  & 853 & 33.8 & 41.1 & 41.1 & 42.9 & \textbf{48.6} & 48.2 & 45.8 & 55.8 & 1.8 & 7.5 & 7.1 & 4.8 & 10.0 \\
GPT-5-mini         & 853 & 39.8 & 48.9 & 48.9 & 50.2 & \textbf{53.3} & 48.4 & 52.8 & 56.3 & 1.3 & 4.4 & -0.5 & 3.9 & 3.5 \\
\bottomrule
\end{tabular}%
}
\caption{Oracle routing and fusion on the 853 answerable MMLongBench-Doc examples. Selectively combining ColBERT and ColPali consistently outperforms either alone, confirming retriever complementarity across all reader models.}
\label{tab:oracle_analysis}
\end{table*}

We test retriever complementarity on 853 answerable MMLongBench-Doc examples under four increasingly favorable retrieval assumptions, with a \textbf{gold ceiling} providing gold pages directly.

Table~\ref{tab:oracle_analysis} reveals three findings. \textbf{Oracle~1} (per-query best-retriever selection) and \textbf{Oracle~2} (answer-level routing that runs both pipelines and keeps the better answer) both improve over the best single retriever, with Oracle~2 yielding larger gains. This confirms that ColBERT--ColPali complementarity is real and that how evidence interacts with downstream reasoning matters as much as retriever quality alone. Strikingly, \textbf{Oracle~3a} (gold-only pages) can \emph{underperform} the ColPali baseline: overly restrictive filtering removes helpful auxiliary context. \textbf{Oracle~3b} (gold pages prioritized, remaining slots filled from ColPali) approaches the gold ceiling most closely ($58.6\%$ vs.\ $59.2\%$ for Qwen3-VL-8B). The bottleneck is learning \emph{which retriever to call at each step and how to compose their outputs}, precisely the routing capability that failure-driven evolution targets directly.

%% file: checklist.tex
\section*{NeurIPS Paper Checklist}

\begin{enumerate}

\item {\bf Claims}
    \item[] Question: Do the main claims made in the abstract and introduction accurately reflect the paper's contributions and scope?
    \item[] Answer: \answerYes{}
    \item[] Justification: The abstract and Introduction summarize the method, scope, and main empirical findings, and the main claims are supported by the Results and Analysis sections.

\item {\bf Limitations}
    \item[] Question: Does the paper discuss the limitations of the work performed by the authors?
    \item[] Answer: \answerYes{}
    \item[] Justification: A dedicated Limitations section discusses single-run evaluation, benchmark scope, computational cost, stopping criterion and stability, and interpretability of evolved policies.

\item {\bf Theory assumptions and proofs}
    \item[] Question: For each theoretical result, does the paper provide the full set of assumptions and a complete (and correct) proof?
    \item[] Answer: \answerNA{}
    \item[] Justification: The paper is empirical and does not contain formal theorems or proofs.

    \item {\bf Experimental result reproducibility}
    \item[] Question: Does the paper fully disclose all the information needed to reproduce the main experimental results of the paper to the extent that it affects the main claims and/or conclusions of the paper (regardless of whether the code and data are provided or not)?
    \item[] Answer: \answerYes{}
    \item[] Justification: Method and Experiments describe the task/meta-agent setup, tools, benchmarks, baselines, metrics, and evaluation protocol used to produce the reported results.

\item {\bf Open access to data and code}
    \item[] Question: Does the paper provide open access to the data and code, with sufficient instructions to faithfully reproduce the main experimental results, as described in supplemental material?
    \item[] Answer: \answerNo{}
    \item[] Justification: The submission does not include an anonymized code release; results are reported on public benchmarks and with cited existing tools and models.

\item {\bf Experimental setting/details}
    \item[] Question: Does the paper specify all the training and test details (e.g., data splits, hyperparameters, how they were chosen, type of optimizer) necessary to understand the results?
    \item[] Answer: \answerYes{}
    \item[] Justification: Experiments specifies the benchmarks, metrics, backbones, retrieval tools, baselines, and checkpoint-selection protocol used in the evaluation.

\item {\bf Experiment statistical significance}
    \item[] Question: Does the paper report error bars suitably and correctly defined or other appropriate information about the statistical significance of the experiments?
    \item[] Answer: \answerNo{}
    \item[] Justification: The paper shows results over multiple iterations highlighting a broader trend of improvements. Further, gains in our study seem to be reasonably high compared to all the baselines our work compares with. 

\item {\bf Experiments compute resources}
    \item[] Question: For each experiment, does the paper provide sufficient information on the computer resources (type of compute workers, memory, time of execution) needed to reproduce the experiments?
    \item[] Answer: \answerNo{}
    \item[] Justification: We have reported frontier models that were used in our experiments. All of the artifacts in our experiments are publicly available, including the frontier models with which we report our scores. 
    
\item {\bf Code of ethics}
    \item[] Question: Does the research conducted in the paper conform, in every respect, with the NeurIPS Code of Ethics \url{https://neurips.cc/public/EthicsGuidelines}?
    \item[] Answer: \answerYes{}
    \item[] Justification: The work is benchmark-based research on document QA and was conducted to the best of our knowledge in accordance with the NeurIPS Code of Ethics.

\item {\bf Broader impacts}
    \item[] Question: Does the paper discuss both potential positive societal impacts and negative societal impacts of the work performed?
    \item[] Answer: \answerYes{}
    \item[] Justification: The Conclusion briefly discusses both beneficial applications of document QA and risks from misuse or over-trust on sensitive documents.
    
\item {\bf Safeguards}
    \item[] Question: Does the paper describe safeguards that have been put in place for responsible release of data or models that have a high risk for misuse (e.g., pre-trained language models, image generators, or scraped datasets)?
    \item[] Answer: \answerNA{}
    \item[] Justification: The paper does not release a new general-purpose foundation model, scraped dataset, or other high-risk asset.

\item {\bf Licenses for existing assets}
    \item[] Question: Are the creators or original owners of assets (e.g., code, data, models), used in the paper, properly credited and are the license and terms of use explicitly mentioned and properly respected?
    \item[] Answer: \answerNo{}
    \item[] Justification: The paper credits benchmarks and prior systems via citations, but it does not enumerate license terms for every external asset in the submission.

\item {\bf New assets}
    \item[] Question: Are new assets introduced in the paper well documented and is the documentation provided alongside the assets?
    \item[] Answer: \answerNA{}
    \item[] Justification: The submission does not introduce a new benchmark, dataset, or public model release.

\item {\bf Crowdsourcing and research with human subjects}
    \item[] Question: For crowdsourcing experiments and research with human subjects, does the paper include the full text of instructions given to participants and screenshots, if applicable, as well as details about compensation (if any)? 
    \item[] Answer: \answerNA{}
    \item[] Justification: The work does not involve crowdsourcing or other human-subject data collection.

\item {\bf Institutional review board (IRB) approvals or equivalent for research with human subjects}
    \item[] Question: Does the paper describe potential risks incurred by study participants, whether such risks were disclosed to the subjects, and whether Institutional Review Board (IRB) approvals (or an equivalent approval/review based on the requirements of your country or institution) were obtained?
    \item[] Answer: \answerNA{}
    \item[] Justification: The work does not involve human subjects research requiring IRB review.

\item {\bf Declaration of LLM usage}
    \item[] Question: Does the paper describe the usage of LLMs if it is an important, original, or non-standard component of the core methods in this research? Note that if the LLM is used only for writing, editing, or formatting purposes and does \emph{not} impact the core methodology, scientific rigor, or originality of the research, declaration is not required.
    \item[] Answer: \answerYes{}
    \item[] Justification: LLMs are core components of both the task agent and the meta-agent, and their role is described in Method and Experiments.

\end{enumerate}